\documentclass[11pt,a4paper]{article}
\usepackage[hyperref]{conll2018}
\usepackage{times}
\usepackage{latexsym}
\usepackage{graphicx}
\usepackage{booktabs}

\usepackage{url}

\usepackage{amsmath}
\usepackage{mathdots}
\usepackage{tikz}
\usetikzlibrary{trees}
\usepackage{tkz-graph} 
\usetikzlibrary{positioning,shapes.geometric, fit, matrix, chains, decorations.pathreplacing, arrows, arrows.meta,quotes,shapes.misc}
\usepackage{pgfplots}

\definecolor{unimittelblau100}{cmyk}{1,.4,0,0}
\definecolor{unimittelblau70}{cmyk}{.7,.2,0,0}
\definecolor{unimittelblau50}{cmyk}{.5,.1,0,0}
\definecolor{unimittelblau30}{cmyk}{.3,.05,0,0}
\definecolor{unianthrazit100}{cmyk}{.5,.2,.2,.85}
\definecolor{unianthrazit70}{cmyk}{.35,.14,.14,.6}
\definecolor{unianthrazit50}{cmyk}{.25,.1,.1,.43}
\definecolor{unianthrazit30}{cmyk}{.15,.06,.06,.26}
\definecolor{univiolettblau100}{cmyk}{.9,.6,0,0}
\definecolor{univiolettblau70}{cmyk}{.68,.4,0,0}
\definecolor{univiolettblau50}{cmyk}{.45,.26,0,0}
\definecolor{univiolettblau30}{cmyk}{.27,.12,0,0}
\definecolor{unigelb}{cmyk}{0,.1,1,0}
\definecolor{uniorange}{cmyk}{0,.7,1,0}
\definecolor{unirot}{cmyk}{0,1,1,0}
\definecolor{unipink}{cmyk}{0,1,0,0}
\definecolor{univiolett}{cmyk}{.6,1,0,0}
\definecolor{unituerkis}{cmyk}{1,0,.4,0}
\definecolor{uniapfelgruen}{cmyk}{.5,0,1,0}
\definecolor{middlegreen}{cmyk}{.6,0,1,0}
\colorlet{phdBoxColor}{unimittelblau30!40}
\colorlet{avgPoolingColor}{uniorange}
\colorlet{wghPoolingColor}{unimittelblau100}
\colorlet{baselineColor}{unianthrazit30!70}
\colorlet{bestColor}{unirot}

\aclfinalcopy 

\title{Comparing Attention-based Convolutional and Recurrent Neural Networks: Success and Limitations in Machine Reading Comprehension}

\author{
  Matthias Blohm, Glorianna Jagfeld, Ekta Sood, Xiang Yu, Ngoc Thang Vu\\
  Institute for Natural Language Processing (IMS) \\
  Universit\"at Stuttgart, Germany \\
  {\tt \{blohmms,jagfelga,soodea,xiangyu,thangvu\}}\\
  {\tt @ims.uni-stuttgart.de}\\}

\date{}

\begin{document}
\maketitle
\begin{abstract}
We propose a machine reading comprehension model based on the compare-aggregate framework with two-staged attention that achieves state-of-the-art results on the MovieQA question answering dataset.
To investigate the limitations of our model as well as the behavioral difference between convolutional and recurrent neural networks, we generate adversarial examples to confuse the model and compare to human performance.
Furthermore, we assess the generalizability of our model by analyzing its differences to human inference, drawing upon insights from cognitive science. 
\end{abstract}

\section{Introduction}
Current state-of-the-art deep learning (DL) models outperform other techniques in many tasks including computer vision~\cite{krizhevsky2012imagenet}, speech recognition~\cite{hinton2012deep} and more recently natural language processing (NLP)~\cite{Collobert:2011tk}. Neural-based NLP systems often use word embeddings~\citep{NNLanguageModel_Bengio03,Embeddings_Collobert08,word2vec_Mikolov13} which are then fed into a convolutional neural network~(CNN)~\citep{CNNImage_Cun90,CNNAudio_Waibel90} or a recurrent neural network~(RNN)~\citep{ElmanNN_Elman90,HochreiterLSTM_97} for further classification.
These approaches proved to be successful for many NLP tasks~\cite{mikolov2010recurrent, CNN, hu2014convolutional, Bahdanau:2014vz}.
Along with the success of DL in a wide range of applications, adversarial examples~\cite{good} - 
that aim to confuse the system 
- have gained popularity in a wide range of research communities such as computer vision and NLP, since they can reveal the limitations in the generalizability of the models.
As opposed to adversarial examples in computer vision, which are computed on continuous data and can thus easily be imperceptible if desired, adversarial attacks in NLP entail the necessity to perform discrete and perceptible changes to the data.
Thus, attack methods for computer vision such as the Fast Gradient Sign Method~(FGSM)~\cite{good} cannot be directly applied to NLP.

Machine comprehension has recently received increased interest in the NLP community~\citep{WikiQA, MQA, SQuAD_Srajpurkar16, chen2016thorough}.
Neural network models perform reasonably well on many data sets with different question answering setups, e.g. multiple choice or answer generation~\citep{CAM, ACN, QANet}. 

Among others, \citet{CAM} proposed the compare-aggregate framework, which uses an attention mechanism~\citep{Attention_Luong15} to compare the question and candidate answers, and a CNN to aggregate information. 
However, there is still an ongoing debate whether CNNs or RNNs are more suitable to NLP~\cite{yin2017comparative}, and the behavioral differences between them are still under research.
Many papers reported remarkable gains when combining these two models in ensembles~\citep{deng2014ensemble, zhou2015c, vu2016combining}, since they process information in different ways and thus are complimentary to each other.

Despite the seemingly high accuracies of many models on machine comprehension tasks, \citet{ADV} argued that many questions in such datasets are easily solvable by superficial cues.
They showed with adversarial examples that most models can be easily tricked by modifications on the data which do not confuse humans.
Similarly, \citet{sanchez2018behavior} performed controlled experiments on the robustness of several Natural Language Inference models by altering hypernym, hyponym, and antonym relations in the data.
Both studies revealed a major weakness of the models:
They largely rely on pattern matching instead of human decision-making processes as required in the tasks, including heuristics~\citep{gigerenzer2011heuristic} and elimination by aspects~\citep{tversky1972elimination}.

In this paper, we implement two machine comprehension models based on the compare-aggregate framework with a hierarchical attention structure using CNNs and RNNs.
First we show that we achieve state-of-the-art results on the MovieQA multiple choice question answering dataset~\citep{MQA} outperforming other systems by a large margin.\footnote{See MovieQA leaderboard, \url{http://movieqa.cs.toronto.edu/leaderboard/}}
Second, we investigate the different behavior of the two systems applying adversarial attacks in a systematic way. To our best knowledge, this is the first work exploring the difference between CNNs and RNNs by such an approach.
Third, we present a detailed comparison between human and machine reading comprehension, giving insights when and why our systems fail.
Therefore, these insights are important for future research towards enhancing machine comprehension systems loosely inspired by human processing.
All code necessary to reproduce our experimental results is made available.\footnote{\url{https://github.com/DigitalPhonetics/reading-comprehension}}

\section{Hierarchical Attention-based Compare-Aggregate Model}
The basis for our model is the compare-aggregate model with attention~\citep{CAM} that has been shown effective for reading comprehension.
We extend the model in two aspects that lead to significant improvements.

Given a preprocessed matrix-representation of the question~$Q$, a text (movie plot)~$P$, and $k$~answer candidates~$A_1 \dots A_k$, the main idea of~\citet{CAM}'s compare-aggregate model is to compare~$P$ to~$Q$ and each~$A_j$ and then aggregate this information into a vector to derive a confidence score~$c_j$ for each answer candidate.

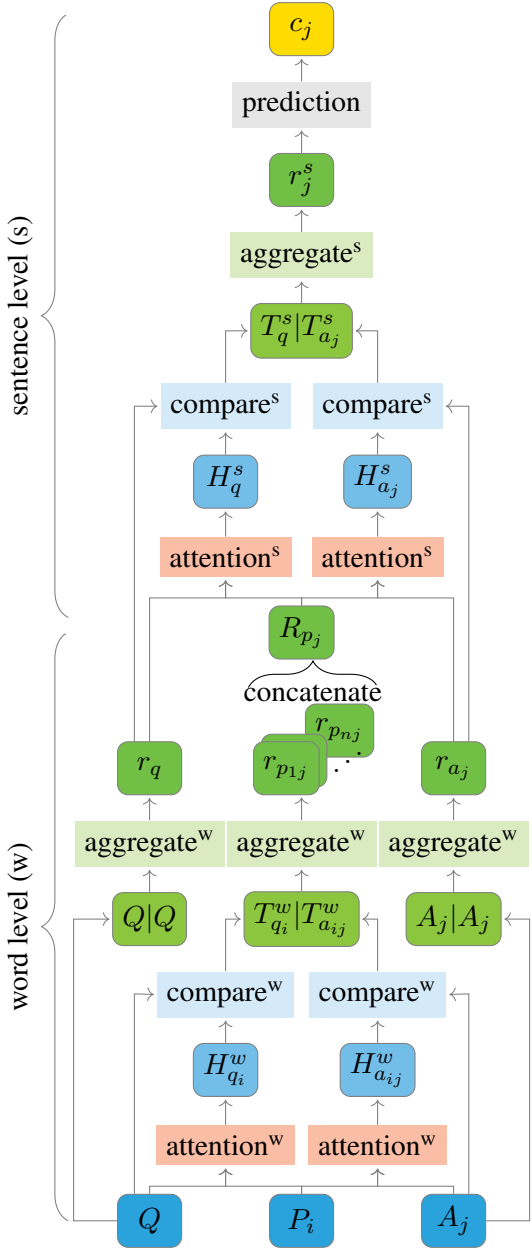
\begin{figure}[h!tb]
	\def\layersep{2.5cm}
	\centering
\begin{tikzpicture}[shorten >=1pt,->,draw=black!50, node distance=\layersep,transform shape]
    \tikzstyle{var}=[rectangle, rounded corners, fill=green!50, draw, minimum height=20pt, minimum width=24pt]
    \tikzstyle{var-input}=[var, fill=unimittelblau70]
    \tikzstyle{var-attention}=[var, fill=unimittelblau50]
    \tikzstyle{var-compared}=[var, fill=uniapfelgruen]
    \tikzstyle{var-aggregated}=[var, fill=middlegreen]
    \tikzstyle{var-confidence}=[var, fill=unigelb]
    \tikzstyle{op}=[rectangle, fill=black!10, minimum height=15pt, minimum width=24pt];
    \tikzstyle{op-attention}=[fill=unirot!30];
    \tikzstyle{op-compare}=[fill=unimittelblau30!50];
    \tikzstyle{op-aggregate}=[fill=uniapfelgruen!30];
    \tikzstyle{edge}=[];
    
    \node[var-input] (Q) at (0,0) {$Q$};
    \node[var-input] (P) at (2,0) {$P_i$};
    \node[var-input] (A) at (4,0) {$A_j$};
    
    \node[op-attention] (att-w-qp) at (1,1) {attention$^\text{w}$};
    \node[op-attention] (att-w-pa) at (3,1) {attention$^\text{w}$};
    
    \node[var-attention] (H-w-q) at (1,2) {$H^w_{q_i}$};
    \node[var-attention] (H-w-a) at (3,2) {$H^w_{a_{ij}}$};
    
    \node[op-compare] (comp-w-qp) at (1,3) {compare$^\text{w}$};
    \node[op-compare] (comp-w-pa) at (3,3) {compare$^\text{w}$};
    
    \node[var-compared] (Q-conc) at (0,4) {$Q|Q$};
    \node[var-compared] (T) at (2,4) {$T^w_{q_i}|T^w_{a_{ij}}$};
    \node[var-compared] (A-conc) at (4,4) {$A_j|A_j$};
    
    \node[op-aggregate] (agg-q) at (0,5) {aggregate$^\text{w}$}; 
    \node[op-aggregate] (agg-t) at (2,5) {aggregate$^\text{w}$};
    \node[op-aggregate] (agg-a) at (4,5) {aggregate$^\text{w}$};
    
    \node[var-aggregated] (r-q) at (0,6) {$r_q$};
    \node[var-aggregated] (r-a) at (4,6) {$r_{a_j}$};
    \node[var-aggregated] (r-p-n) at (2.5,6.5) {$r_{p_{nj}}$};
    \node[var-aggregated] (r-p-2) at (1.9,6.1) {$r_{p_{2j}}$};
    \node[var-aggregated] (r-p-1) at (1.8,6) {$r_{p_{1j}}$};
    \node[rectangle] (dots) at (2.6,6.2) {$\iddots$};
    
    \draw [decorate,black,-,decoration={brace,amplitude=10pt,raise=2pt},yshift=0pt]
    (1.3,7) -- (3.0,7) node [black,midway,yshift=0pt] {concatenate};
    \node[var-aggregated] (r-p) at (2,7.8) {$R_{p_j}$};
    
    \draw[edge] (Q.north) -- +(0,0.1) -| (att-w-qp.south);
    \draw[edge] ([xshift=-2mm]Q.north) |- +(0,2.65) -- (comp-w-qp.west);
    \draw[edge] (P.north) -- +(0,0.1) -| (att-w-qp.south);
    \draw[edge] (P.north) -- +(0,0.1) -| (att-w-pa.south);
    \draw[edge] (A.north) -- +(0,0.1) -| (att-w-pa.south);
    \draw[edge] ([xshift=2mm]A.north) |- +(0,2.65) -- (comp-w-pa.east);
    
    \draw[edge] (att-w-qp) -- (H-w-q);
    \draw[edge] (att-w-pa) -- (H-w-a);
    \draw[edge] (H-w-q) -- (comp-w-qp);
    \draw[edge] (H-w-a) -- (comp-w-pa);
    
    \draw[edge] (comp-w-qp) |- (T);
    \draw[edge] (comp-w-pa) |- (T);
    
    \draw[edge] (Q) -- +(-1,0) |- (Q-conc);
    \draw[edge] (A) -- +(1,0) |- (A-conc);
    
    \draw[edge] (Q-conc) -- (agg-q);
	\draw[edge] (T) -- (agg-t);    
    \draw[edge] (A-conc) -- (agg-a);
    
    \draw[edge] (agg-q) -- (r-q);
    \draw[edge] (agg-t) -- +(0,0.65);
    \draw[edge] (agg-a) -- (r-a);

    \def \secondlayerstart {8.8}
    
    \node[op-attention] (att-s-qp) at (1,\secondlayerstart) {attention$^\text{s}$};
    \node[op-attention] (att-s-pa) at (3,\secondlayerstart) {attention$^\text{s}$};
    
    \node[var-attention] (H-s-q) at (1,\secondlayerstart+1) {$H^s_{q}$};
    \node[var-attention] (H-s-a) at (3,\secondlayerstart+1) {$H^s_{a_{j}}$};
    
    \node[op-compare] (comp-s-qp) at (1,\secondlayerstart+2) {compare$^\text{s}$};
    \node[op-compare] (comp-s-pa) at (3,\secondlayerstart+2) {compare$^\text{s}$};
    
    \node[var-compared] (T-s) at (2,\secondlayerstart+3) {$T^s_q|T^s_{a_j}$};
    \node[op-aggregate] (agg-s) at (2,\secondlayerstart+4) {aggregate$^\text{s}$};
    \node[var-aggregated] (r-s) at (2,\secondlayerstart+5) {$r^s_{j}$};
    \node[op] (dense) at (2,\secondlayerstart+6) {prediction};
    \node[var-confidence] (c) at (2,\secondlayerstart+7) {$c_{j}$};

    \draw[edge] (r-p.north) -- +(0,0.1) -| (att-s-qp.south);
    \draw[edge] (r-p.north) -- +(0,0.1) -| (att-s-pa.south);
    \draw[edge] (r-q.north) -- +(0,1.9) -| (att-s-qp.south);
    \draw[edge] ([xshift=-.2cm]r-q.north) -- +(0,1.9) |- (comp-s-qp.west);
    \draw[edge] (r-a.north) -- +(0,1.9) -| (att-s-pa.south);
    \draw[edge] ([xshift=2mm]r-a.north) -- +(0,1.9) |- (comp-s-pa.east);
    
    \draw[edge] (att-s-qp) -- (H-s-q);
    \draw[edge] (att-s-pa) -- (H-s-a);
    \draw[edge] (H-s-q) -- (comp-s-qp);
    \draw[edge] (H-s-a) -- (comp-s-pa);
    
   	\draw[edge] (comp-s-qp) |- (T-s);
    \draw[edge] (comp-s-pa) |- (T-s);
    
    \draw[edge] (T-s) -- (agg-s);
    \draw[edge] (agg-s) -- (r-s);
    \draw[edge] (r-s) -- (dense);
    \draw[edge] (dense) -- (c);
    
    \draw [decorate,-,decoration={brace,amplitude=10pt,raise=2pt},yshift=0pt]
    (-1.0,0.0) -- (-1.0,7.8) node [black,midway,xshift=-20pt] {\rotatebox{90}{word level (w)}};
    \draw [decorate,-,decoration={brace,amplitude=10pt,raise=2pt},yshift=0pt]
    (-1.0,8.0) -- (-1.0,16) node [black,midway,xshift=-20pt] {\rotatebox{90}{sentence level (s)}};

%
%
%
%
%
\end{tikzpicture}
	\caption{Hierarchical compare-aggregate model to compute the confidence score~$c_j$ of a preprocessed answer candidate~$A_j$ given question~$Q$ and plot~$P=P_1 \dots P_n$.}
	\label{fig:model}
\end{figure}

\citet{CAM} concatenate all plot sentences and do not leverage the inherent structuring of the text into sentences.
Inspired by the recent success of hierarchical models in NLP~\citep{HierarchicalIR_Sordoni15,HierarchicalDocClass_Yang16,ACN}
we extend the model to perform comparison and aggregation on the word and sentence level separately (see Figure~\ref{fig:model}).
Specifically, we first apply the compare-aggregate model to each plot sentence~$P_i$ individually to obtain question and answer-weighted representations~$T^w_{q_i}$, $T^w_{a_{ij}}$ for each sentence.
We then run the aggregation operation on each sentence representation individually to obtain sentence vector representations $r_{p_{ij}}$.
The sentence representations are concatenated to obtain a plot representation~$r_{p_{j}}$, which enters the sentence level of comparison and aggregation that mirrors the word level architecture.

As a second modification of the base model, we implement an RNN-based aggregation function to replace the CNN-based aggregation originally proposed by~\citet{CAM}.
In the following we detail the building blocks of our hierarchical attention-based compare-aggregate model as depicted in Figure~\ref{fig:model}.

\textbf{Preprocessing}
We represent the words in the question~$q$, the plot sentences~$p_i$ and the answer candidates~$a_j$ by pretrained embeddings to obtain matrices $\overline{Q}, \overline{P}, \overline{A_j}$.
We project them to lower dimensional $Q, P, A_j$ via the following operation:

\begin{equation}
X = \sigma\left(W^i\overline{X} + b^i \right) \odot \tanh\left(W^u\overline{X} + b^u \right)
\end{equation}

\textbf{Attention}
The attention operation weights the plot regarding the question or a candidate answer.

\begin{align}
G &= \text{softmax}\left (X^{T}P \right)\\
H &= XG,
\end{align}

where $X$ on the word level represents $Q$ or an answer candidate $A_j$  and on the sentence level $r_{q}$ or $r_{a_j}$.\footnote{Different from~\citet{CAM} we use dot-product attention instead of general attention~\citep{Attention_Luong15} because we found no benefit of the additional parameters of general attention in preliminary experiments.}

\textbf{Comparison}
The comparison operation performs an element-wise comparison of each~$h_l$ in~$H$ with its counterparts~$q_l/a_{j_l}$ on the word level and $r_{q}/r_{a_{j}}$ on the sentence level, respectively.
\citet{CAM} compared many comparison functions.
Here we use only the \textsc{SUBMULT} function since it performed best for MovieQA:

\begin{equation*}t_l = \text{ReLU}(W\left[\begin{array}{c}(x_l - h_l) \odot (x_l - h_l)\\ x_l \odot h_l \end{array}  \right] + b)
\end{equation*}

where $\odot$~denotes element-wise multiplication and $x_l$ corresponds to entries of $Q/A_j$ or $r_q/r_{a_j}$.

\textbf{Aggregation}
The goal of the aggregation operation is to condense the information of a variable-length sequence into a single vector.
\citet{CAM} implemented the aggregation operation as a single-layer CNN following~\citet{CNN}.
Specifically, they used a 1D~convolution with filter sizes~\{1,3,5\}, to capture unigrams, trigrams and 5-grams.

\begin{equation}
	\text{aggregate}_{\text{CNN}} = \text{CNN}([z_1 \dots z_m])
\end{equation}

where~$[z_1 \dots z_m]$ on the word level corresponds to the sequence of row vectors of~$Q, T^w = T^w_{q_i}|T^w_{a_{ij}}, A_j$, and on the sentence level to that of~$T^s = T^s_q|T^s_{a_j}$. 

While CNNs are effective in modeling location-independent n-gram patterns, they cannot capture longer-range dependencies.
Yet, we argue that it is important to also consider the context of the matched phrases.
This motivates our proposed sequential aggregation function based on a single-layer unidirectional RNN with Long Short-Term Memory~(LSTM) units~\citep{HochreiterLSTM_97}.

\begin{equation}
\text{aggregate}_{\text{RNN-LSTM}} = \text{RNN}([z_1 \dots z_m])
\end{equation}

By performing 1-max pooling over the outputs of $\text{aggregate}_{\text{CNN}}$ or $\text{aggregate}_{\text{RNN-LSTM}}$\footnote{Using only the last RNN output for $\text{aggregate}_{\text{RNN-LSTM}}$ did not provide convincing results.} we obtain a single vector $r$ (representing~$r_q, r_{p_{ij}}, r_{a_j}$ on the word level, or $r^s_j$ on the sentence level):

\begin{equation}
r = \text{max\_pool(aggregate}([z_1 \dots z_m]))
\end{equation}

We share the weights between the comparison and aggregation operations within the word and sentence level but not across levels.

\textbf{Prediction}
We map each aggregated answer-specific plot representation~$r^s_j$ to a confidence score~$c_j$ by two dense layers with shared weights for all answer candidates and of which the first uses $\tanh$ activation and the second one no activation function.
The confidence scores are normalized to form a probability distribution~$p_1 \dots p_k$ by a softmax operation.

\section{Experimental Set-Up}

The hyperparameters for our models are provided in \S\ref{sec:hyperparams} in the appendix.

\subsection{Data}
We evaluate our models on the MovieQA dataset~\cite{MQA} that contains 14,944~multiple-choice questions on 408~movies collected by human annotators.
The questions vary from simple ``{\it who}" or ``{\it when}" to more complex ``{\it why}" or ``{\it how}" question types.
Each question is provided along with five candidate answers of which only one is correct.  

While the dataset contains multiple sources of information about the movie contents such as videos, subtitles, and movie scripts, here we focus on answering the questions only from \emph{plot synopses}.
Plot synopses are summaries of the movies collected from Wikipedia that mostly describe the actions happening in the story.
They were used as references for the question collection and so far yield the best results on the dataset according to the MovieQA leaderboard.
Figure~\ref{fig:exMov} shows a sample question together with its candidate answers and an excerpt of the corresponding movie plot which contains the necessary information to answer the question.
The dataset is split into 9,848 training, 1,958 development and 3,138 test questions, respectively.
Note that the test set accuracies can only be evaluated by submitting the predictions to the server.

\begin{figure}[tbh]
\begin{tabular}{l}
	\textbf{Plot:} \dots Aragorn is crowned King of Gondor\\
	and taking Arwen as his queen before all\\
	present at his coronation bowing before Frodo\\
	and the other Hobbits. The Hobbits return to\\\textbf{the Shire} where Sam marries Rosie Cotton. \dots\\
	\midrule
	\textbf{Question:} Where does Sam marry Rosie? \\\midrule
	\textbf{Candidate Answers:} 0) Grey Havens\\
	1) Gondor 2) \textbf{The Shire} 3) Erebor 4) Mordor
\end{tabular}%
\caption{\emph{MovieQA} example question \citep{CAM}.}
\label{fig:exMov}
\end{figure}

\section{Results}
\label{sec:results}
We train 11~models with different random initializations for both the CNN and RNN-LSTM aggregation function and form majority-vote ensembles of the nine models with the highest validation accuracy.
Table~\ref{tab:rel} shows the accuracies of ensembles of our proposed model variations in comparison to the published results on the MovieQA validation and test set.
To the best of our knowledge, the results of~\citet{CAM} and \citet{LRS} were achieved by single models, while the results of~\citet{ACN} corresponds to an ensemble of multiple models.

All our hierarchical single and ensemble models outperform the previous state of the art on both the validation and test set.
With a test accuracy of~85.12, the RNN-LSTM ensemble achieves a new state of the art that is more than five percentage points above the previous best result.

The hierarchical structure is crucial for the model's success.
Adding it to the CNN that operates only at word level\footnote{The CNN word level only model essentially corresponds to our reimplementation of~\citet{CAM}. The performance gain on the validation set might be due to using consistent random initializations for unknown words.} causes a pronounced improvement on the validation set.

\begin{table}[tb]
	\centering
	\begin{tabular}{lrr}  
		Systems          & Val.         & Test      \\\midrule
		\textit{\citet{CAM}}  & 72.10         & 72.90             \\ 
		\textit{\citet{ACN}} & 79.00        & 79.99             \\ 	 
		\textit{\citet{LRS}} & -        		& 80.02          \\
		\midrule \multicolumn{3}{c}{Proposed models}\\ \midrule
		CNN word level only & 76.51     	& 		-		\\ 
		\textbf{CNN} & 79.62      	& -			\\ 
		\textbf{CNN ensemble} & 82.58      	& 82.73			\\
		\textbf{RNN-LSTM} & 83.14	      	& - 				\\ 
		\textbf{RNN-LSTM ensemble} & 84.37	      	& \textbf{85.12} 				\\\midrule	 
		\textbf{CNN RNN-LSTM ensemble} &  \textbf{84.78}      	& 84.70 \\ 
		
	\end{tabular}
	\caption{MovieQA accuracies for previously published results and our proposed single models (best out of~11) and ensembles (nine best out of~11).}  
	\label{tab:rel} 
\end{table}

Furthermore, the RNN-LSTM aggregation function is superior to aggregation via CNNs, improving the validation accuracy by 1.5~percentage points.
While this improvement is statistically significant,\footnote{McNemar test~\cite{McNemar47},~$p < 0.05$.} combining both aggregation functions by ensembling the nine best CNN and RNN-LSTM models each, yields a small but statistically insignificant improvement of 0.41~percentage points over the RNN-LSTM ensemble on the validation set.
This might explain why the RNN-LSTM ensemble even outperforms the CNN RNN-LSTM ensemble on the test set by a small margin.
The difference in test set performance between these two ensembles is likely not significant.
We cannot test this as the test set is not released and only accuracy values can be obtained for model evaluation on the test set.

\subsection{Impact of Sentence Attention}

The sentence attention allows us to get more insight into the models' inner state.
For example, it allows us to check whether the model actually focuses on relevant sentences in order to answer the questions.
The MovieQA dataset provides human annotations of the minimal set of plot sentences required to answer a question.
In average, 1.15/1.11 sentences in the training/validation set are marked as containing the clue to the answer.
We leverage this information and compute the ranks of these relevant plot sentences according to the models' sentence attention distribution.
We extract the plot sentence relevance scores after the sentence-level comparison operation as average of $T^s_q$ and $T^s_{a_j}$, where $a_j$ corresponds to the selected answer of the model.
As Table~\ref{tab:results-attention} reveals, both model variants pay most attention to the relevant plot sentences for 70\%~of the cases.
Identifying the relevant sentences is an important success factor: Relevant sentences are ranked highest only in 35\%~of the incorrectly solved questions.

\begin{table}[]
	\centering
	\begin{tabular}{lrr}
		Systems          & CNN        & RNN-LSTM     \\\midrule
		All questions &  71.45 &	71.31	\\ 
		- Correctly solved &  80.86 & 	79.35		\\	 
		- Incorrectly solved & 35.73 &	34.49	\\
	\end{tabular}
	\caption{Percentage of questions in which the plot sentences containing the clues for the answer are ranked highest according to the model's sentence attention distribution (relative to its selected answer) on the validation set (averaged results of nine models).}  
	\label{tab:results-attention} 
\end{table}

\section{Limitations}
To help us identifying the models' weaknesses, we design a series of systematic adversarial attacks. 
These attacks are defined in different categories depending on the linguistic level (word vs. sentence level) and the knowledge of the adversaries (black-box vs. white-box).
According to the taxonomy proposed by~\citet{yuan2017adversarial}, black-box and white-box attacks differ in the access of the adversary to the trained neural network model. In black-box settings, the adversary acts as a standard user that has only access to the output of the model in form of labels or confidence scores. In contrast, the adversary in white-box settings has access to all the details of the models such as training data, network architectures and hyperparameters.
In this work, the white-box adversary has access to the attention weights of the model at the word and sentence level.
We apply all our attacks to the nine selected models~(see~\S\ref{sec:results}) for each aggregation type.

\subsection{Word-level Black-box Attack}
Adversarial examples for image recognition are typically created by adding some imperceptible noise~\citep{VisualAdvEx_Szegedy14, VisualAdvEx_Goodfellow15}, yet this is difficult to do for natural language  because of its discrete nature.
The closest analogue would be paraphrasing but high-quality paraphrases are difficult to obtain automatically:
Recent attempts with a sophisticated paraphrase-generation system based on a large paraphrase database yielded about 20\%~contradicting adversarial examples~\citep{AdversarialExampleGenerationSyntacticParaphrase_Iyyer18}.

Thus, we designed an adversarial black-box attack on the questions based on manual lexical substitution.
We inspected the 106~most frequent words of the validation set questions and manually defined lexical substitutions of single words and multiword expressions of up to two tokens wherever applicable.
We made sure that the lexical substitutions were meaning preserving and resulted in grammatical sentences in all contexts.\footnote{We only substituted with words contained in the pretrained GloVe embeddings used by the models to avoid introducing unknown words. Even though we did not restrict the substitutes to words from the training set vocabulary, it turned out that all selected words and multiword expressions were indeed contained in the training set vocabulary already, except for the synonym \emph{buddy} for \emph{friend}.}
Our final set of 51~substitution rules resulted in a modification of 25\%~of the validation set questions.

\begin{table}[]
	\centering
	\begin{tabular}{lrr}
		Systems          & Average       & Ensemble      \\\midrule
		CNN &  78.74   	& 	81.72			\\ 
		RNN-LSTM &   81.53 	& 83.76 			\\	 
		CNN RNN-LSTM &  81.14  	& 84.27			\\	
	\end{tabular}
	\caption{Adversarial accuracies on the validation set under the word-level black-box attack based on manual lexical substitutions in questions.}  
	\label{tab:results-adv-word-black} 
\end{table}

As can be seen from Table~\ref{tab:results-adv-word-black}, the models are quite robust against meaning-preserving lexical substitutions: The accuracy drops by less than one percentage point for all ensembles.
Although the differences are small, the RNN-LSTM and CNN RNN-LSTM ensembles are even less affected by lexical substitutions than the CNN ensemble.
By only modifying the questions, we have likely reduced their lexical overlap with the answer candidates and the plots.
The robustness of the models against this attack can probably be attributed to the pretrained GloVe embeddings, which allow it to generalize for semantically equivalent lexical choices.
Stronger attacks involving substitutions with more infrequent words that do not appear in the pretrained embeddings could show the limitation of the models in this respect.
We leave the automatic generation of further-reaching adversarial examples based on paraphrases to future work.

\subsection{Word-level White-box Attack}
We performed a word-level white-box adversarial attack in which we used the models' internal attention distributions to explicitly target the plot words they base their decision on.
More precisely, in this experiment we leveraged the models' sentence-level attention distribution to find the plot sentence it gave most weight to conditioned on the correct answer.
In this sentence, the~$k$ words that received most attention were then exchanged by randomly chosen words from the MovieQA vocabulary.

\begin{figure}[tb]
	\begin{tikzpicture}
\begin{axis}[
    axis lines*=middle,
    xmin=0, xmax=7,
    ymin=0, ymax=105,
    xtick={1,2,3,4,5,6,7},
    ytick={20,40,60,80,100},,
    width=\linewidth,
    xticklabels={1,2,3,5,10,20,40},
    x label style={at={(axis description cs:0.5,-0.01)}},
    y label style={at={(axis description cs:0.05,0.5)}},
    ylabel = accuracy (\%),
    xlabel = replaced words k,
    xtick distance=0.5,
   legend style ={ at={(1,1)}, 
   	anchor=north east, draw=black, 
   	fill=white,align=left},
    smooth
]
\addplot [color=unimittelblau100, mark=triangle*, dashdotted] table {
0			100
1			100
2			95
3			80
4			75
5			40
6			25
7			20
};
\addlegendentry{human};
\addlegendentry{CNN};
\addlegendentry{RNN-LSTM};
\addlegendentry{chance};
\addplot [color=unirot, densely dashed, mark=*] table {
0	79.16
1	66.12
2	53.43
3	38.9
4	35.73
5	31.73
6	31.59
7	31.62
};
\addplot [color=uniapfelgruen, mark=diamond*] table {
0	82.07
1	68.95
2	57.84
3	43.78
4	39.76
5	33.82
6	32.66
7	32.6
};
\addplot [color=unianthrazit100, densely dotted] coordinates {(0,20) (7,20)}; 
\end{axis}
\end{tikzpicture}
	\caption{Adversarial accuracies on the validation set under the word-level white-box attack based on word exchange. $k$~is the number of words that are modified in the plot sentence with most attention (average accuracies over nine models).
	Human evaluation is based on 20~randomly sampled questions with plots attacked for a single CNN~model (single annotator, one of the authors of this paper).}  
	\label{fig:results-adv-word-white} 
\end{figure}
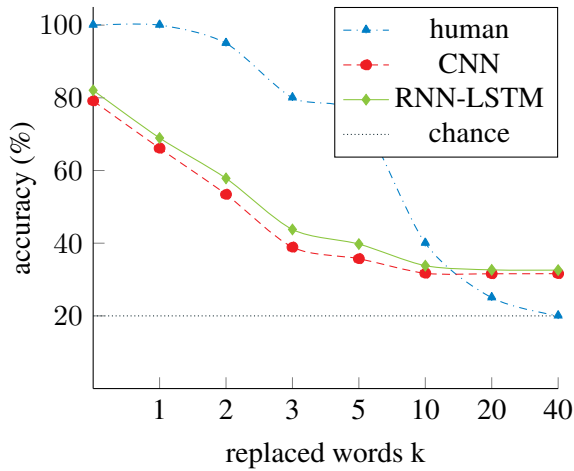

As Figure~\ref{fig:results-adv-word-white} reveals, already modifying the single most important word in the most important sentence has a large effect on the average performance of both the CNN and RNN-LSTM models.
For increasing~$k$, the RNN-LSTM versions appeared to be a bit more robust against the attack, but for~$k \geq 10$ the difference shrinks and the accuracy of both models drops to only about~30\%.
This experiment shows that manipulating the  most relevant plot information by removing important words makes the model fail quickly, since it is no longer able to draw correct conclusions for the questions without the necessary plot context.
Although the human annotator proved more robust against this attack for a small number of replaced words, increasing~$k$ beyond five showed the same drastic decline in performance.

\subsection{Sentence-level Black-box Attacks}
In order to find out to which extent our models are susceptible to distracting information added to the plot, we adapt the \emph{AddAny} attack by~\citet{ADV} originally designed for the SQuAD reading comprehension dataset. 
This adversarial attack consists of adding a distractor sentence $s$ at the end of the plot, regardless of grammaticality.
The word sequence $s = w_1 w_2 \dots w_{10}$ is initialized by ten common English words.
Then each word is greedily changed from a pool of 20~random common words (\emph{AddC}) to minimize the model's confidence score for the correct answer.
We refer the reader to~\citet{ADV} for the full details of this attack.
Likewise we generate adversarial sentences using a pool of ten random common words for each $w_i$ in conjunction with all question words (\emph{AddQ}) or additionally the words from all incorrect answer candidates (\emph{AddQA}).
While these attacks do not take any particular measures to prevent the added sentence from contradicting the correct answer, this is very unlikely given the ungrammatical nature of the generated word sequences.

\begin{table}[]
	\centering
	\def \width {0.8cm}
	\begin{tabular}{p{1.99cm}p{\width}p{\width}p{\width}p{\width}}  
		Systems		& Orig.  & \emph{AddC}   & \emph{AddQ}  & \emph{AddQA}           \\\midrule
		 \multicolumn{5}{c}{Without optimization}\\\midrule
		CNN 		& 76.87 &	76.67 & 76.66 & 76.33 \\ 
		RNN-LSTM 	& 81.11 & 	81.11 & 81.05 & 81.05	\\\midrule
		 \multicolumn{5}{c}{After two optimization epochs}\\\midrule
		CNN 		& N/A 	&	73.38 & 57.39 & 13.61 \\ 
		RNN-LSTM  	& N/A 	&	79.94 & 68.05 & 23.22	\\	
	\end{tabular}
	\caption{Adversarial accuracies on 200~random validation questions under the sentence-level black-box attacks (averaged results of nine models).} 
	\label{tab:results-adv-sent-black} 
\end{table}

The first two rows in Table~\ref{tab:results-adv-sent-black} show the effect of appending a random sentence to the plot.\footnote{As this attack is computationally very expensive we only ran it on a random subset of 200~validation questions for two optimization epochs of the distractor sentence.}
The impact on performance is fairly small indicating the robustness of both models.
However, after only two epochs of optimizing the selected words in the added sentence, the performance drops markedly under all variants of the sentence-level black-box attacks as displayed in the two bottom rows of Table~\ref{tab:results-adv-sent-black}.
While composing the sentence of just common English words~(\emph{AddC}) does not affect the models too much, adding words from the question and incorrect answers~(\emph{AddQA}) is most detrimental and causes both models to perform at or even below chance level.
The models' performance under \textit{AddQ}, where the distractor sentence does not contain answer candidate words, is much higher than under \textit{AddQA}.
We observe that the models can be easily distracted by adding a single sequence of significant words, even though it bears no semantic relation to the rest of the plot.
This suggests that both models heavily rely on the content of the provided answer candidates and might just perform matching of learned patterns to select the right answer.

Another observation is that the RNN-LSTM models outperform the CNN models by a large margin under all attacks.
The stronger the attack, the larger is the performance gap, indicating that RNNs depend less on pattern matching and are less prone to this kind of attack.
Figure~\ref{fig:attW_b4} and~\ref{fig:attW_after} in the appendix provide an example of the sentence and word attention distributions of a CNN model before and after the \textit{AddQA} attack.

\begin{table}[]
	\centering
	\begin{tabular}{lrrr}
						& \multicolumn{2}{c}{Attack optimized for}\\
		Evaluated systems        & CNN  & RNN-LSTM             \\ \midrule
		CNN & 13.61 & 	21.50 	\\ 
		RNN-LSTM & 22.06 & 23.22		\\
		
	\end{tabular}
		\caption{
			\emph{AddQA} attack results when testing models on adversarial examples optimized to fool another model (averaged results of nine models).}  
	\label{tab:results-adv-sent-black_2} 
\end{table}

To test the transferability of the adversarial examples across models, we test the CNN models on the adversarial examples optimized to fool the RNN models and vice versa.
As Table~\ref{tab:results-adv-sent-black_2} shows, the performance of both models is degraded to the same level independent of the model the attack was optimized for.
This suggests that both models suffer from similar weaknesses.

A straightforward way to try to improve the models' robustness against adversarial attacks is to mix some adversarial examples into the training data.
\citet{ADV} evaluated this for the \textit{AddAny} attack on SQuAD and found that training on a mix of adversarial and original samples indeed improves the performance with respect to this specific adversarial attack.
Yet a slight change of the attack, e.g. adding the distracting sentence as first instead of last sentence, made the adversarially-trained models to fail almost as badly as without adversarial training.
Therefore, we argue that it is more promising to look for general improvements of the model than training on adversarial examples generated by a specific attack.

\subsection{Sentence-level White-box Attack}
Instead of modifying the words in the sentence we also attempted to attack the model by removing the whole plot sentence with the highest attention.
In this experiment, we wanted to test (1) if the model really focuses on the most important sentence, so it would become more difficult to answer the question, and (2) if the model is able to pick up more subtle cues or perform answer elimination to still be able to infer the correct answer with some confidence.
As can be seen from Table~\ref{tab:results-adv-sent-white}, the accuracy decreases dramatically for both models by removing only one plot sentence.
This proves that the model indeed focuses on the correct sentence where the hint to answer a question is given.
These results correspond to those of the white-box attack at word level with a large number \emph{k} of modified words. 
For the remaining 30\% of correctly answered questions we observed that sometimes the models still were able to answer correctly because of the context information provided in other plot sentences.

We also measured human performance under this attack on 20~randomly sampled questions on distinct plots, where the sentence containing the answer information was removed.
A single annotator (one of the authors of this paper) achieved~55\% accuracy on this task, which is way above chance level and the models' performance.
The human reported to be able to answer nine questions with reasonable confidence by deducing from other information distributed across the plots; two answers were correct by guessing.
Answering the questions under this attack took a lot of time and effort.
This highlights the weakness of the model to give answers in more complex scenarios where the answer is less obvious.
\begin{table}[]
	\centering
	\begin{tabular}{lrr}
		Systems          & Average        & Ensemble      \\\midrule
		CNN &  31.59 & 	32.07			\\ 
		RNN-LSTM & 32.61 & 32.17 			\\		
	\end{tabular}
	\caption{Adversarial accuracies on the validation set under the sentence-level white-box attack based on removal of the plot sentence with highest attention (averaged results of nine models).
	}  
	\label{tab:results-adv-sent-white} 
\end{table}

\section{Human vs. Machine Processing}
In order to gain insights how to further improve machine reading comprehension, we performed a case study in which a human was asked to answer difficult questions that none of 11~CNN or RNN-LSTM models solved correctly.
The human evaluator obtained the plots and the questions with the corresponding five answer candidates; having access to the information in the same manner as the models. 
There are clear motifs in the type of reasoning and logic required, inherent to human cognition.
In this light, we aim at inferring the gap between the model's and human cognitive information processing to identify problems followed by potential solutions. 

Since we were especially interested in getting insights on human strategies for the cases where our models failed, 50 difficult questions of the CNN models in the validation set were analyzed by a human evaluator. 
All of the questions were correctly answered by the human evaluator noticing several key postulates:  textual entailment, choice by elimination, referential knowledge and their combination~\citep{hummel2005relational}.

Textual entailment is required to solve 60\% of the questions, such as the question ``{\it What do Matt, Steve, and Andrew record themselves doing weeks after their experience in the woods?}" with the relevant sentence ``{\it Weeks later, Andrew, Matt, and Steve record themselves as they display telekinetic abilities, but begin bleeding from their noses when they overexert themselves}".
The human predicts the correct answer, ``{\it Moving objects with their mind}", based on world knowledge of the word {\it telekinetic}.
A further example in this regard is the question ``{\it Do the robbers take people in the bank as hostage?}" with the relevant sentence ``{\it They seize control of a Manhattan bank and take the employees and patrons hostage.}"
The human picks the correct answer ``{\it Yes, they do}", as \emph{people in the bank} is a hypernym of \emph{employees} and \emph{patrons} in this context.
Lacking notion of these semantic relations, the model answers incorrectly.

The process of elimination and heuristics proved essential to solve 44\% of the questions. One example is ``{\it Where is New Penzance located?}" with the relevant sentence ``{\it In September 1965, on a New England island called New Penzance, 12-year-old orphan Sam Shakusky is attending Camp Ivanhoe [\dots]}".
The human could not infer the answer ``{\it Off the coast of North Carolina}" from reading the plot alone, as this region is not inherently known to be associated with New England, the location mentioned in the plot.
However, by using the process of elimination and heuristics, the annotator was able to deduce the likely answer
with the certainty that the other candidates are less likely correct.
Additionally, with the ranking of keywords, humans can infer the correct answer in examples such as the question ``{\it What kind of classes does Toula take up?}", with the relevant sentence ``{\it After some persuasion by his wife, Maria [\dots], Gus reluctantly permits Toula to begin taking computer classes at a local community college
[\dots]}".
In this case, the human identified the keywords {\it classes} and {\it Toula}.
The word {\it classes} obtains a higher ranking as it appears in three of the five possible answers.
Ultimately, the correct prediction was made using ranking and the main keyword to find the correct answer, ``{\it Computer classes}".

Referential knowledge is presumed in 36\% of the questions, e.g. in the question ``{\it What does Stigman do with the money?}" with the relevant sentence ``{\it After the heist, Stigman follows orders to betray Trench and escape with the money, managing to pull his gun right as Trench is about to pull his own}".
The human chooses the correct answer ``{\it He takes it}", however the models select either ``{\it He splits it with Trench}" or ``{\it He leaves it in the vault}". When analyzing the plot, we can see that the two pronouns, {\it He} and {\it it}, are ambiguous to the models but clear to the human, leading to incorrect model predictions.
The variance is due to the notion that humans have the ability to understand the referents from the plot. Another example where lack of referential knowledge effects the models' performances, but not the human, can be observed with the question ``{\it What happens to any human who is encountered in Narnia?}" with the relevant sentence ``{\it If a human is encountered they are to be brought to her}". The human is able to select the correct answer, ``{\it They are to be brought to the White Witch}", even though the plot refers to the character by the pronoun {\it her}. 

Furthermore, it is apparent that many questions expect a combination of various reasoning skills.
The question ``{\it What is Xavier's mutant ability?}" with the relevant sentence ``{\it Present are Lehnsherr, now known as Magneto, and the telepathic Professor Charles Xavier, who privately discuss their differing views on the relationship between humans and mutants}", depicts this phenomena.
The human reports that she utilized the keywords {\it Xavier}, {\it mutant} and {\it ability}, raking {\it Xavier} more predominantly.
By identifying {\it Professor Charles Xavier} in the plot as referent of the most important keyword, she could eliminate the  incorrect answers.

The human evaluator also conducted an extensive comparison of the baseline word-level models with the hierarchical CNN models.
In particular, she looked at those questions where the performance of both model types differed most (in terms of the number of models out of 11 that solved the question correctly).
There are 101 validation questions which the majority of hierarchical CNN models solved correctly but only a minority (at least six less) word-level models did so.
These were compared to the 28 validation questions on which the word-level models outperformed the hierarchical ones.

No prevailing pattern could be identified for the few instances where the word-level models did better than the hierarchical ones.
Yet, we found some evidence that the hierarchical models seem to do better for questions requiring matching longer answer candidates and handling lexical variation.
An example for such a more complex question is ``{\it What happens to Deon in the end?}". The relevant plot sentence is ``{\it He then transfers the dying Deon's consciousness into a spare robot through the modified MOOSE helmet}", and the correct answer ``{\it His consciousness is transferred into a robot}".
All answer candidates consist of at least five words; the lexical overlap between the question and correct answer with the plot sentence is just \{\textit{into, a, robot}\}.
While only two baseline models identify the correct answer, all but one of the hierarchical models do so.

Additionally, among the 101 questions where the hierarchical models do far better than the word-level models there are only very few (18) questions where none of the word-level models predicted the correct answer.
It seems to be the case that the hierarchical structure helps the model to gain confidence, causing more models to make the correct prediction. 
An example for this is the question ``{\it What does Lucius tell Harry?}”, where the relevant sentence is ``{\it Lucius reveals that Harry only saw a dream of Sirius being tortured; it was a method to lure Harry into the Death Eaters' grasp, not an actual situation.}”, and the correct answer is ``{\it His vision of Sirius being tortured was a dream used to lure Harry to the Death }”.
The majority of the word-level models predicted an incorrect answer ``{\it His vision of Sirius being tortured was true}”, and only five of them selected the correct answer.
In contrast, all hierarchical models solved this question correctly.

The same comparison was conducted between the hierarchical CNN and RNN-LSTM models.
Although there are improvements, which indicate that sequential processing is better suited for QA tasks, the RNN-LSTM models exhibit the same fundamental drawbacks.
They suffer from co-reference errors, lack the entailment ability, and are inefficient at keyword elimination. 
This observation reveals the fundamental weaknesses of our proposed network architecture and indicates directions for future improvements.

\section{Conclusion}
We proposed a machine reading comprehension model based on the compare-aggregate framework with a hierarchical attention structure that achieves state-of-the-art results on the MovieQA question answering dataset, greatly outperforming previous models. 
Then, we explored the limitations of our models and the behavioral difference between CNNs and RNN-LSTMs with adversarial examples generated at different linguistic levels~(word vs. sentence level) and from different adversary's knowledge (black-box vs. white-box). 
In general, RNN-LSTM models outperformed CNN models, but our results for sentence-level black-box attacks indicate they might share the same weaknesses.

Finally, our intensive analysis on the differences between the model and human inference suggest that both models seem to learn matching patterns to select the right answer rather than performing plausible inferences as humans do.
The results of these studies also imply that other human like processing mechanism such as referential relations, implicit real world knowledge, i.e.,  entailment, and answer by elimination via ranking plausibility~\citep{hummel2005relational} should be integrated in the system to further advance machine reading comprehension.

\bibliography{../thesis/references}
\bibliographystyle{acl_natbib_nourl}

\appendix

\cleardoublepage

\section{Supplemental Material}
\label{sec:supplemental}

\subsection{Model Details}
\label{sec:hyperparams}

The word embeddings are initialized from 300-dimensional pretrained GloVe vectors~\citep{GLO}\footnote{\url{http://nlp.stanford.edu/data/glove.840B.300d.zip}.} that are kept fixed during training.
Words not contained in these embeddings are initialized randomly, while all updatable weights are initialized according to the Xavier initialization~\citep{XAV}.

We train all models for five epochs, evaluating their performance on the validation set after each epoch.
We keep the model with the best validation accuracy, which is usually achieved after one, or sometimes two or three training epochs.

For the CNN models, we started with the hyperparameters of~\citet{CAM} and tuned the dropout rate of the pretrained embeddings, learning rate, and L2~regularization weight on the validation set.
For the RNN-LSTM model we started from the hierarchical CNN model's hyperparameters and tuned the dropout and learning rate again. 
Table~\ref{tab:hyperparameters} lists the hyperparameters for our models.

\begin{table}[h]
	\centering
	\begin{tabular}{l|ll}
		& \textbf{CNN}    & \textbf{RNN-LSTM}  \\
		\hline
	embedding size & 300 &	300\\
	weight initialization    & xavier              & xavier\\
	batch size     & 30                  & 30\\
	optimizer      & Adam                & Adam \\
	dropout rate        & 0.0                & 0.0\\
	learning rate  & 0.001               & 0.0025\\
	regularization & L2 & L2 \\
	reg. weights & 0.0001 & 0.0001 \\
	size of dense layers & 150 & 150\\
	CNN kernel heights & 	[1,3,5] & N/A \\
	LSTM units & N/A &	150 \\
	\end{tabular}
	\caption{Hyperparameters for the CNN and RNN-LSTM hierarchical attention-based compare-aggregate models.}
	\label{tab:hyperparameters}
\end{table}

\subsection{Adversarial Examples}
\label{sec:visualization}
\begin{figure*}[h!tb]
	\includegraphics[width=\textwidth]{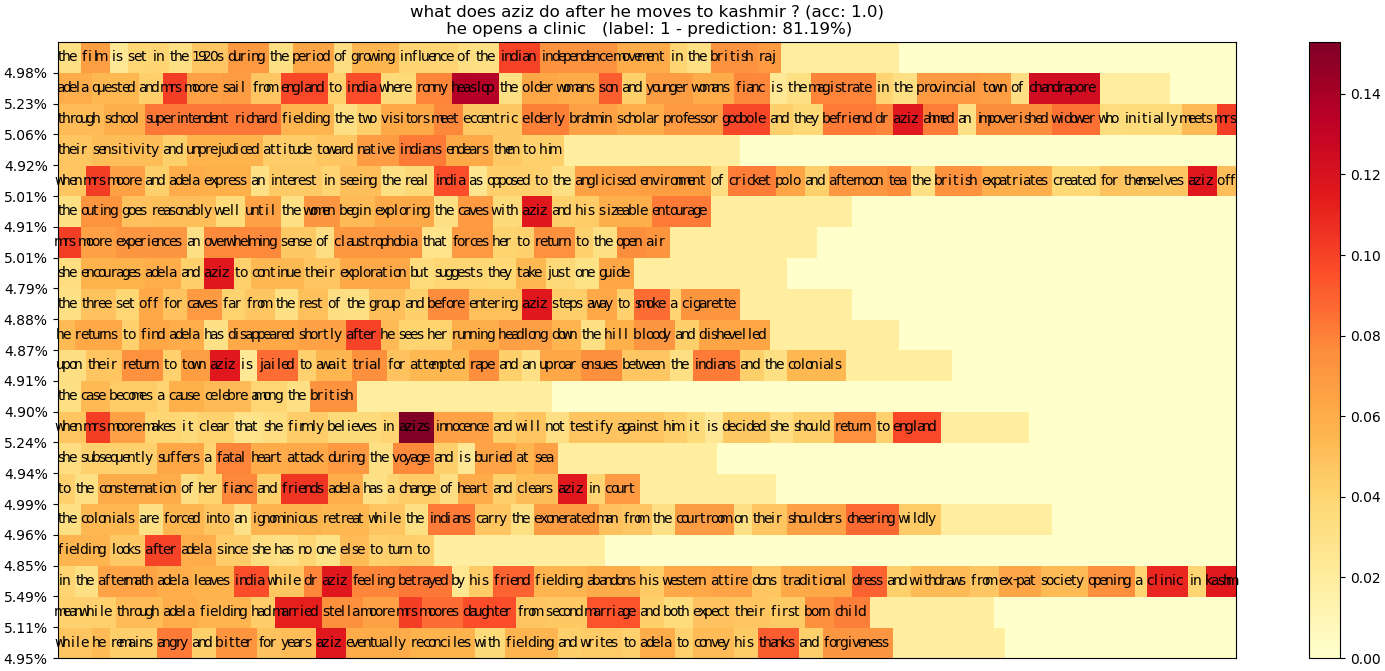}
	\caption{Attention weights visualization of the CNN model.}
	\label{fig:attW_b4}
\end{figure*}

\begin{figure*}[h!tb]
	\includegraphics[width=\textwidth]{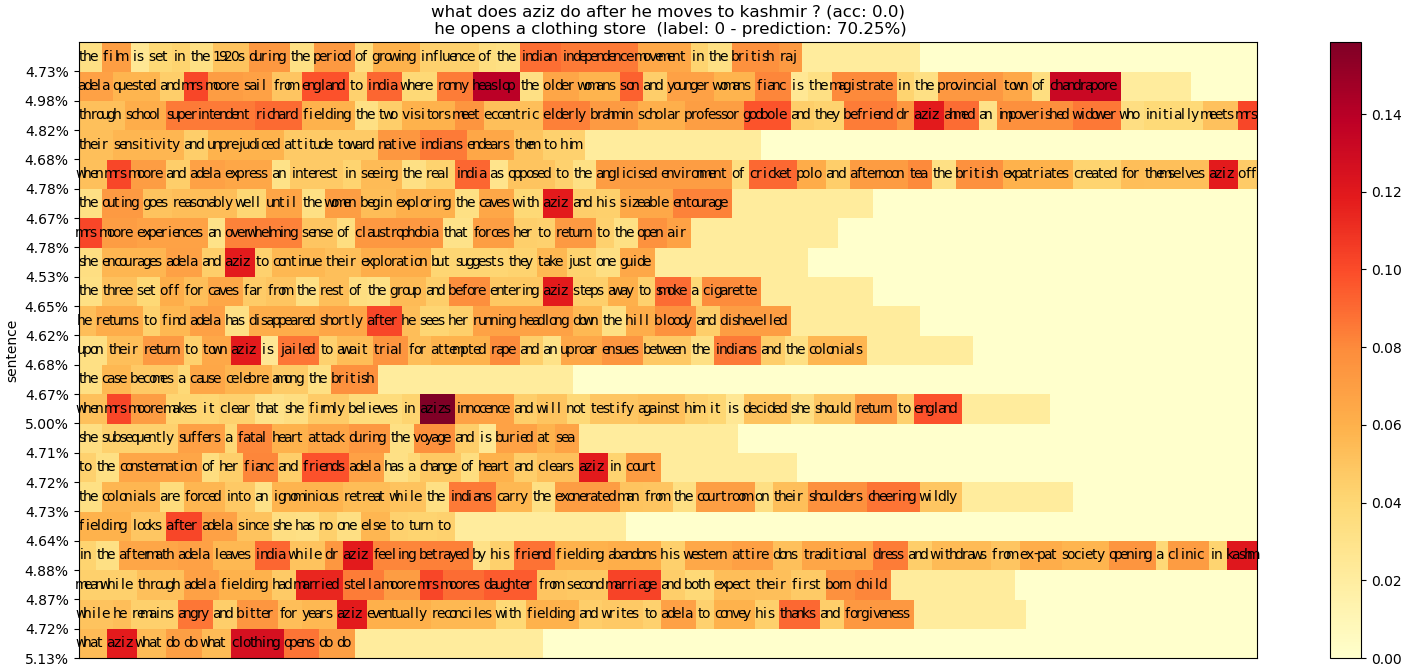}
	\caption{Attention weights visualization of the CNN model after attacking it with \textit{AddQA}, which added the final sentence to the plot.}
	\label{fig:attW_after}
\end{figure*}

Figure~\ref{fig:adEx} shows a MovieQA example question and an adversarial sentence generated by the \textit{AddQA} approach.
The adversarial sentence that is appended to the plot has a very high overlap with the question and one of the wrong answers.
Although the sentence is grammatically wrong and meaningless, our model picked the wrong answer~4) instead of the correct~1). Figure~\ref{fig:attW_b4} and~\ref{fig:attW_after} illustrate the sentence and the word attention weights of a CNN model before and after the adversarial attack.
\begin{figure}[h]
\centering
\begin{tabular}{l}
        \textbf{Adversarial sentence:} what aziz what do \\ do what clothing opens do do \\ \hline
	\textbf{Question:}What does Aziz do after he moves \\ to Kashmir? \\ \hline
	\textbf{Candidate Answers:} \\ 0) He opens a mosque\\
	1) \textbf{He opens a clinic}\\ 2) He opens a school \\ 3) He becomes a monk \\ 4) He opens a clothing store
\end{tabular}%
\caption{\emph{MovieQA} example question and the generated adversarial sentence using \textit{AddQA}.}
\label{fig:adEx}
\end{figure}

\end{document}